\documentclass[letterpaper, 10 pt, conference]{ieeeconf}  %

\IEEEoverridecommandlockouts                              %

\usepackage{graphics} %
\usepackage{epsfig} %
\usepackage{mathptmx} %
\usepackage{times} %
\usepackage{amsmath} %
\usepackage{amssymb}  %
\usepackage{graphicx}
\usepackage{mathtools}
\usepackage{bm}
\usepackage{bbold}
\usepackage{pifont}%
\usepackage[pagebackref,breaklinks,colorlinks]{hyperref}
\usepackage{tabularray}
\usepackage{subcaption}
\usepackage[export]{adjustbox}

\newcommand{\minus}{\scalebox{0.75}[1.0]{$-$}}

\title{\LARGE \bf
EventTransAct: A video transformer-based framework for Event-camera based action recognition
}

\author{Tristan de Blegiers*, Ishan Rajendrakumar Dave*, Adeel Yousaf and Mubarak Shah     %
\thanks{All authors are with Center for Research in Computer Vision lab (CRCV), University of Central Florida, USA. * denotes equal contribution.} 
\noindent \thanks{\tt\{tristandeblegiers, ishandave\}@knights.ucf.edu,
adeel.yousaf@knights.ucf.edu, shah@crcv.ucf.edu}}

\begin{document}

\maketitle
\thispagestyle{plain}
\pagestyle{plain}

\begin{abstract}
Recognizing and comprehending human actions and gestures is a crucial perception requirement for robots to interact with humans and carry out tasks in diverse domains, including service robotics, healthcare, and manufacturing. 
 Event cameras, with their ability to capture fast-moving objects at a high temporal resolution, offer new opportunities compared to standard action recognition in RGB videos. 
However, previous research on event camera action recognition has primarily focused on sensor-specific network architectures and image encoding, which may not be suitable for new sensors and limit the use of recent advancement in transformer-based architectures. 
In this study, we employ using a computationally efficient model, namely the video transformer network (VTN), which initially acquires spatial embeddings per event-frame and then utilizes a temporal self-attention mechanism. This approach separates the spatial and temporal operations, resulting in VTN being more computationally efficient than other video transformers that process spatio-temporal volumes directly.
In order to better adopt the VTN for the sparse and finegrained nature of event data, we design Event-Contrastive Loss ($\mathcal{L}_{EC}$) and event specific augmentations. Proposed $\mathcal{L}_{EC}$ promotes learning fine-grained spatial cues in the spatial backbone of VTN by contrasting temporally misaligned frames. We evaluate our method on real-world action recognition of N-EPIC Kitchens dataset, and achieve state-of-the-art results on both protocols - testing in seen kitchen (\textbf{74.9\%} accuracy) and testing in unseen kitchens (\textbf{42.43\% and 46.66\% Accuracy}). Our approach also takes less computation time compared to competitive prior approaches.  We also evaluate our method on the standard DVS Gesture recognition dataset, achieving a competitive accuracy of \textbf{97.9\%} compared to prior work that uses dedicated architectures and image-encoding for the DVS dataset.
These results demonstrate the potential of our framework \textit{EventTransAct} for real-world applications of event-camera based action recognition. Project Page: \url{https://tristandb8.github.io/EventTransAct_webpage/}

\end{abstract}

\section{INTRODUCTION}

The ability to recognize and interpret human actions and gestures is critical for robots to interact with people and perform tasks in various domains, such as manufacturing, healthcare, and service robotics. Some of such actions could be reaching, grasping, or pointing, as they often provide important cues about the user's intentions and needs.

Advancements in deep neural architectures~\cite{gberta_2021_ICML, aim_chen, videomae, gabv2, rizve2021gabriella} and large-scale datasets~\cite{hvu, kinetics, hacs, miech2019howto100m} have significantly improved the performance of action recognition. Traditionally, action recognition has been performed using video-based sensors, such as RGB cameras. However, these sensors have several limitations in robotics applications. For example, they require high computational resources to process large amounts of data, which can be a challenge for resource-constrained robots. They are also sensitive to motion blur and lighting conditions, which can reduce their action recognition performance and reliability~\cite{schiappa2022large}.

Event-based sensors, on the other hand, offer several advantages for action recognition in robotics \cite{gallego2020event}. These sensors only capture changes in the scene, rather than full frames, which makes them more efficient and robust to high-speed movements and lighting changes. They also provide high temporal resolution, which can capture fast and subtle actions that may be missed by video-based sensors. Additionally, the event cameras also avoid breaching the visual private information of the user such as skin color, gender, clothing, etc, while recognizing the action \cite{ahmad2022event, dave2022spact, fioresi2023tedspad}. Therefore, there is growing interest in developing event-based action recognition methods for robotics applications.

\begin{figure}
    \centering
    \vspace{2mm}
    \begin{subfigure}{0.46\linewidth}
        \centering
        \includegraphics[width=\linewidth]{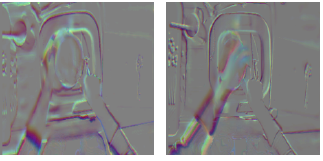}
        \vspace{-6mm}
        \caption{\texttt{wash}}
        \vspace{3mm}
    \end{subfigure}
    \hspace{2mm}
    \begin{subfigure}{0.46\linewidth}
        \centering
        \includegraphics[width=\linewidth]{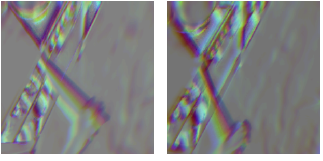}
        \vspace{-6mm}
        \caption{\texttt{close}}
        \vspace{3mm}
    \end{subfigure}
    \begin{subfigure}{0.46\linewidth}
        \centering
        \includegraphics[width=\linewidth]{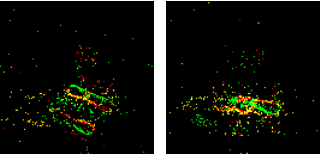}
        \vspace{-6mm}
        \caption{\texttt{arm roll}}
    \end{subfigure}
    \hspace{2mm}
    \begin{subfigure}{0.46\linewidth}
        \centering
        \includegraphics[width=\linewidth]{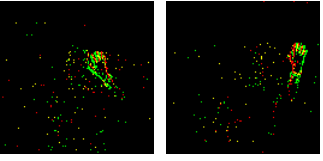}
        \vspace{-6mm}
        \caption{\texttt{left hand wave}}
    \end{subfigure}

    \caption{\textbf{Event-based actions samples} First two rows are from real-world action recognition N-EPIC Kitchens~\cite{nepic} dataset, whereas, second row shows samples from DVS gesture recognition dataset~\cite{dvs}.}
    \label{fig:samples}
    \vspace{-4mm}
\end{figure}
Event-based action recognition has been focused mainly on gesture recognition~\cite{dvs} or egocentric action recognition in the real-world cooking videos~\cite{nepic}. Since these datasets are acquired using different devices and post-processing, they look very different. Some samples and their action labels are shown in Fig~\ref{fig:samples}. The event-specific representation proposed for one dataset may not be effective for the other. For instance, spiking neural architectures are not suitable for N-Epic Kitchens dataset since it is acquired using different a type of sensor. Also, event-based action recognition methods like~\cite{nepic} have not focused on leveraging the advancement of recent neural architectures like vision transformers which have shown great improvement in learning spatio-temporal dependencies of the actions ~\cite{vtn, gberta_2021_ICML, zhang2021vidtr} compared to prior vision architectures including convolutions and aggregation modules like LSTM~\cite{hochreiter1997long}. In this work, our goal is to leverage the recent vision transformer and develop a training framework which can be suitable for different event-based action datasets. 

\begin{figure*}[h]
    \vspace{2mm}
    \centering
    \includegraphics[width=.85\linewidth]{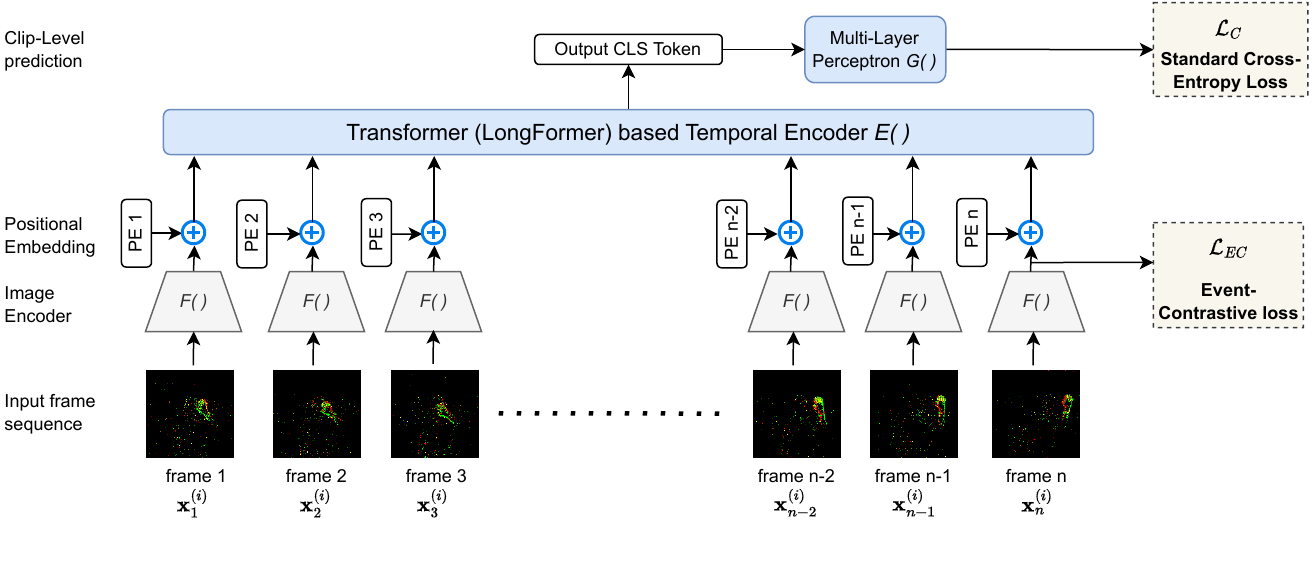}
    \vspace{-3mm}
    \caption{We employed a Video Transformer Network (VTN) architecture~\cite{vtn}, which expands an image encoder $F$ with a transformer $E$ that uses frame embeddings as input tokens. Each event-frame $\mathbf{x}^{(i)}_t$ is passed through a spatial-encoder $F(\cdot)$ to extract spatial features, which are then processed by a transformer-based LongFormer module $E(\cdot)$, augmented with positional embeddings (PE) and a [CLS] token, to learn global temporal dependencies. The [CLS] token output is then classified by head $G(\cdot)$. The model is trained using both cross-entropy loss and a proposed event contrastive loss.      
    Details about network architecture can found in Sec~\ref{subsec:vtn} and loss functions can be found in Sec~\ref{subsec:ecl}.
    }
    \label{fig:framework}
    \vspace{-3mm}
\end{figure*}

Most standard action recognition methods that utilize RGB cameras rely on 3-dimensional operations to capture spatio-temporal changes across frames. However, with event-stream data, a single event-frame is capable of capturing spatio-temporal changes, thus eliminating the need for spatio-temporal volume-based operations. Instead, we employ separate spatial and temporal operations which are more computationally efficient. To achieve this, we leverage the Video Transformer Network (VTN)~\cite{vtn} (Fig~\ref{fig:framework}). VTN computes spatial embeddings for each frame and applies temporal attention to aggregate these embeddings across time.

Event data is typically more sparse and fine-grained compared to RGB data. Therefore, a standard video-level classification loss is not adequate to train the spatial-backbone of VTN for event data. To overcome this limitation, we propose a within-instance contrastive objective called \textit{``Event-Contrastive Loss"}. This loss promotes temporal-distinctiveness across the spatial embeddings of the same video, aiding the spatial backbone to capture fine-grained changes in the event data.

The primary contribution of our work can be summarized as follows:

\begin{itemize}
    \item Our work proposes a video transformer network-based framework for event-camera based action recognition. In this framework, we introduce an event-contrastive loss and specific augmentations to better adapt the network to event data.
    \item Our model is evaluated on the real-world N-EPIC Kitchens~\cite{nepic} dataset for the purpose of classifying actions in kitchen videos. We achieve \textit{state-of-the-art} results in both testing protocols—within seen kitchens and unseen kitchens. Importantly, our method offers a lower computational cost compared to previous approaches.
    \item We further assess our method's performance on the standard DVS~\cite{dvs} Gesture recognition dataset, achieving an impressive \textbf{97.9\%} accuracy. This result is competitive with prior works which employ dedicated architectures and image-encoding specifically for the DVS dataset.
\end{itemize}

\section{Related Work}
\noindent\textbf{Event-based Vision}
In recent years event cameras have emerged as a valuable vision modality for various computer vision applications, ranging from low-level tasks such as object tracking \cite{Jiao_2021_CVPR}, detection \cite{perot2020learning} and optical flow estimation \cite{8880496} to high-level tasks such as image reconstruction \cite{stoffregen2020reducing}, recognition \cite{kim2021n}, and segmentation \cite{alonso2019ev}.  These applications benefit from the unique properties of event cameras such as their ability to capture fine-grained temporal resolution and process visual information in an efficient manner.

Research in the field of event cameras can be broadly categorized into two main groups \cite{chen2022ecsnet}. The first group includes methods that treat events as an asynchronous stream in conjunction with spiking neural networks (SNN) \cite{lu2020event}. The second group comprises techniques that initially convert events to an image-like representation, which allows for an extension of the existing literature \cite{Zhu_2019_CVPR, rebecq2019events}.
 Although sparse event data is suitable for SNNs, the absence of dedicated hardware and effective back-propagation algorithms makes it difficult to utilize and optimize SNNs \cite{ahmad2022event}.
On the other hand, image-like representations can take advantage of the existing deep-learning literature. 

\noindent\textbf{Action Recognition}
 is a popular topic in the computer vision community, deep learning-based approaches succeed without needing tedious feature engineering \cite{kinetics}.

While there is a large body of work on action recognition~\cite{ulhaq2022vision} employing RGB videos, the research on event-based action recognition is still in its infancy \cite{sun2022human}. Intrinsic motion information encoded by event data motivated~\cite{dvs} to exploit event cameras for gesture recognition, a subset problem of action recognition. Specifically designed event-based neuromorphic processor TrueNorth is used to achieve gesture recognition in real-time. A recent work~\cite{nepic} shows that the environmental bias in egocentric action recognition can be solved using event cameras. CNN-based architectural variations are introduced to handle the inter-channel relations in the event data. Moreover, ~\cite{nepic} uses optical flow to distill motion information in event data in a student-teacher approach. The key challenge to processing the event data is to effectively extract the fine-grained temporal information from asynchronous and sparse data.

The existing literature on event-based vision is often limited to utilizing specific network architectures based on the acquisition sensor, and has not fully explored the potential of recent developments in RGB data. However, in recent years, vision transformers have demonstrated the ability to enhance spatio-temporal reasoning in videos through self-attention mechanisms~\cite{gberta_2021_ICML, zhang2021vidtr, vtn}. In this paper, we propose a computationally efficient video transformer network solution that can perform both gesture and action recognition on various event datasets.

\noindent\textbf{Contrastive Learning} has become a popular method for training neural networks without labeled data~\cite{simclr,moco, swav}, as it enables unsupervised learning and promotes learning of fine-grained features through instance-discrimination objectives~\cite{wu2018unsupervised}. In recent years, several works have applied intra-instance contrastive losses to videos to increase temporal distinctiveness in the features of the same video~\cite{tclr, timebalance}. Our aim is to adopt the spatial-backbone of VTN model to the sparse and finegrained event data. To achieve this, we propose a frame-level intra-instance contrastive loss to improve the spatial embeddings of the VTN by encouraging frame-to-frame differences and enhancing temporal distinctiveness.

\section{Method}
\label{sec:method}

In this section, we describe a novel framework for event-camera-based action recognition. In Sec~\ref{subsec:formulation}, we first formulate the problem. In Sec.~\ref{subsec:vtn},  we explain our network architecture in detail, whereas in Sec.~\ref{subsec:ecl}, we provide details on proposed event-contrastive loss. 
\subsection{Problem Formulation}
\label{subsec:formulation}
Event-based action recognition is a simple multi-class classification problem, where an input video $\mathbb{V}^{(i)}$ is classified into one of the $N_C$ possible action classes. After learning on a training set $D_{train}$, the learned model $\theta$ is evaluated on a disjoint test set $D_{test}$. We measure the performance of the model in terms of top-1 classification accuracy.

\subsection{Video Transformer Network}
\label{subsec:vtn}
Event data is aggregated into frame-encoding following prior work~\cite{nepic} to get a video $\mathbb{V}^{(i)}$ from the event sequence $i$. Let $\mathbf{y}^{(i)}_{c} \in [0, N_C]$ be the action-class label of $\mathbb{V}^{(i)}$. Each video has $T$ frames, from which we randomly sample $n$ frames and define it as clip $\mathbb{X}^{(i)}= \{\mathbf{x}^{(i)}_1, \mathbf{x}^{(i)}_2, ..., \mathbf{x}^{(i)}_n\}$, where $\mathbf{x}^{(i)}_t$ represents an event-frame at time $t$ and $n<T$. 

We utilize a video transformer network (VTN)~\cite{vtn} to learn from training data.  Unlike other video architectures like C3D~\cite{c3d}, I3D~\cite{kinetics}, X3D~\cite{feichtenhofer2020x3d} or 3D-ResNet~\cite{kenshohara} which have spatiotemporal learnable modules, VTN provides a computationally efficient way by introducing separable spatial and temporal modules. A schematic of our framework is shown in Fig~\ref{fig:framework}. VTN architecture consists mainly of three components: Spatial Encoder $F(\cdot)$, Temporal Encoder $E(\cdot)$, and classification head $G(\cdot)$. 

First of all, each event-frame $\mathbf{x}^{(i)}_t$ (where $t \in [1,n]$)  is passed through spatial-encoder $F(\cdot)$ to capture spatial features. These spatial features are responsible for capturing the shape and appearance of individual event-frames. 

Now, the goal is to learn the temporal features like object motion, pose-changes, gesture,  etc. from the spatial features. To learn global temporal dependencies of the spatial features, a transformer-based LongFormer module $E(\cdot)$ is used. Instead of self-attention~\cite{vaswani2017attention} over the full sequence, LongFormer implements a constraint attention window which reduces the computation. Each spatial feature is appended with a positional embedding (PE) to associate its relative position in time before passing through $E(\cdot)$ as an input token. A special classification token ([CLS]) is also attached to the sequence of spatial features. Finally, the output [CLS] token is passed through the classification head $G(\cdot)$ to achieve the final prediction.

The VTN model is trained using standard cross-entropy loss and proposed event contrastive loss. The supervision from the ground-truth label $\mathbf{y}^{(i)}_{c}$ is utilized using standard cross-entropy loss, as depicted in the following equation:

\begin{equation}\label{eq:crossentropy}
\mathcal{L}^{(i)}_{C} = -\sum_{c=1}^{N_C} \mathbf{y}^{(i)}_{c}\log \mathbf{p}^{(i)}_{c}
\end{equation}

\noindent For instance $i$, $\mathbf{p}^{(i)}_{c}$ is the classwise prediction vector. 

\subsection{Event Contrastive Loss}
\label{subsec:ecl}

\begin{figure*}[h]
    \centering
    \vspace{2mm}
    \includegraphics[width=.85\linewidth]{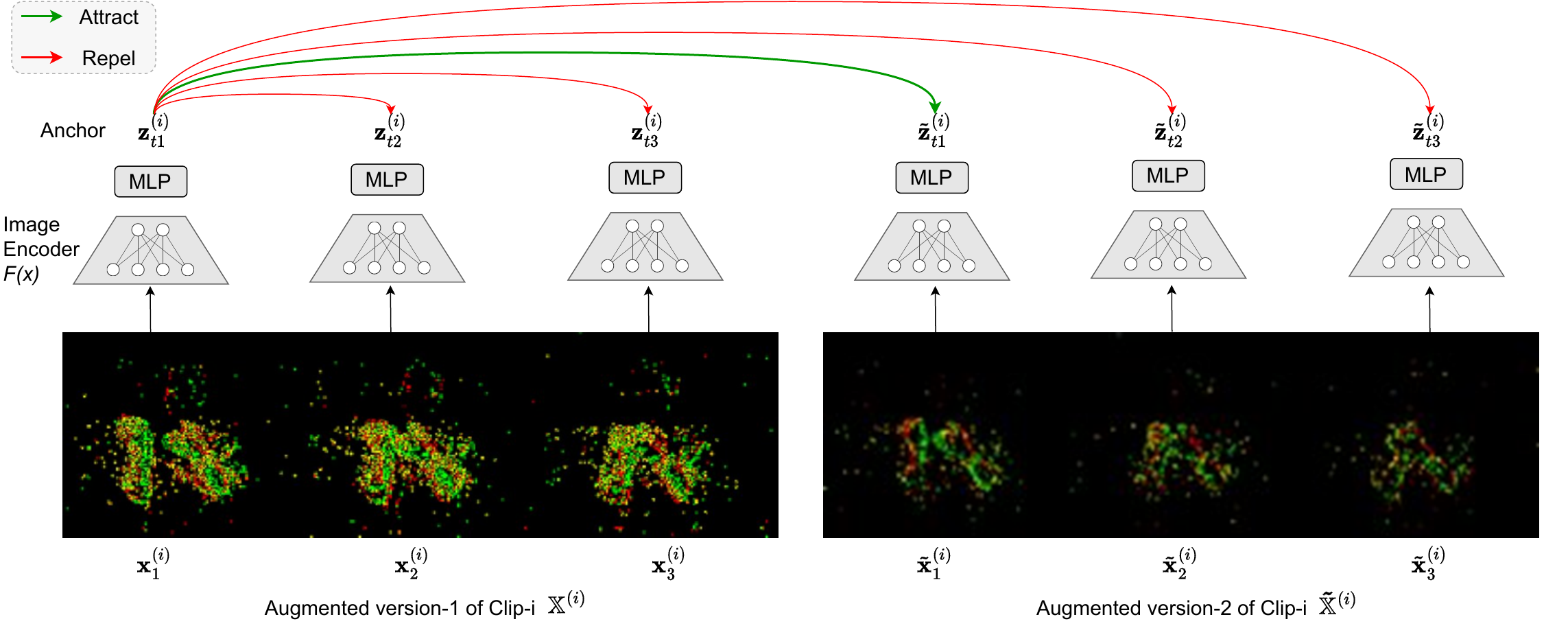}
    \caption{\textbf{Event Contrastive Learning} increases temporal-distinctiveness of the spatial embedding $F(x)$ by maximizing the agreement between two differently augmented version of the same frame, whereas maximizes the disagreement between \textit{temporally misaligned} frames. For visualization purpose only 3 frames per clip are shown. Details in Section~\ref{subsec:ecl}.}
    \label{fig:ecl}
    \vspace{-3mm}
\end{figure*}
Standard Cross-Entropy loss focuses on learning the features at the \textit{video-level} and may not effectively utilize the capacity of the spatial encoder and may perform suboptimally due to overfitting with the lower amount of data. In order to learn the \textit{fine-grained spatial features}, we derive event-contrastive loss from the standard contrastive loss formulation. 

Contrastive loss deals with maximizing the agreement between anchor and positive whereas maximizing disagreement between anchor and negatives. Our proposed event contrastive learning  deals with learning temporally-distinctive features across the event-frames of the same video instance $i$. 
In order to accomplish this, a new clip denoted as $\mathbb{\tilde{X}}^{(i)} = \{\mathbf{\tilde{x}}^{(i)}_{t}\}_{t=1}^{n}$ is created, which comprises randomly augmented versions of event-frames present in $\mathbb{X}^{(i)}$. These randomly augmented clips are shown in Fig~\ref{fig:ecl}. The embeddings of spatial encoder are projected through a non-linear projection head $P(\cdot)$. The projection of each event-frame can be defined as $\mathbf{z}^{(i)} = P(F(\mathbf{x}^{(i)}))$.

Next, we maximize the agreement between the projections of a pair of frames $\{(\mathbf{z}^{(i)}_{t_1},\mathbf{\tilde{z}}^{(i)}_{t_1}) \mid t_1 \in \{1..n\} \}$ from the same timestamp and maximize the disagreement between the projections of pair of temporally misaligned frames. A mathematical expression for this contrastive objective can be written as:
\vspace{-1mm}
\begin{equation}\label{eq:dist}
  \mathcal{L}_{EC}^{(i)}=- \sum_{t_1=1}^{n} \log \frac{\mathrm{h}\left(\mathbf{z}^{(i)}_{t_1}, \mathbf{\tilde{z}}^{(i)}_{t_1}\right)}{\sum\limits_{\substack{t_2=1 \\ t2\neq t_1}}^{n} \mathrm{h}(\mathbf{z}^{(i)}_{t_1}, \mathbf{z}^{(i)}_{t_2}) + \mathrm{h}(\mathbf{z}^{(i)}_{t_1}, \mathbf{\tilde{z}}^{(i)}_{t_2})]},
\end{equation}
\normalsize	
\noindent where $\mathrm{h}(\mathbf{u_{1}}, \mathbf{u_{2}})=\exp \left(\mathbf{u_{1}}^{T}\mathbf{u_{2}}/(\|\mathbf{u_{1}}\| \|\mathbf{u_{2}}\| \tau) \right)$ is used to compute the similarity between $\mathbf{u_{1}}$ and $\mathbf{u_{2}}$ vectors with an adjustable temperature parameter, $\tau$. $\mathbb{1}_{[j\neq i]} \in \{0, 1\}$ is an indicator function which equals 1 iff $j \neq i$.

Finally, to train the entire framework we add both $\mathcal{L}^{(i)}_{C}$ (Eq.~\ref{eq:crossentropy}) and $\mathcal{L}_{EC}^{(i)}$ (Eq.~\ref{eq:dist}) as shown in the following equation: 

\begin{equation}\label{eq:total}
\mathcal{L}^{(i)}= \mathcal{L}^{(i)}_{C} + \alpha \mathcal{L}^{(i)}_{EC}, 
\end{equation}
\noindent where $\alpha$ is a weighting factor between two losses.

\subsection{Event specific augmentations}
\label{subsec:aug}
Augmentations play a crucial role in contrastive learning as shown in multiple prior contrastive learning works~\cite{simclr, tclr, mocov3}. If we do not have meaningful augmentations, then contrastive loss may not help in learning meaningful representation for the downstream task (here, action recognition). Note that if we use simple augmentations like cropping, scaling, etc. then $\mathcal{L}_{EC}$ becomes trivial to solve. To this extent, we look into more specific augmentations for the event-based camera. We encourage the reader to watch the supplementary video while reading this section. In order to get the event-frame from the real event data, we first need to aggregate the events with a temporal window (denoted as $\rho$) to generate a frame. Employing a high $\rho$ results in more events per frame, therefore, the fine-grained spatial structure is lost, as shown in augmented version-1 in Fig~\ref{fig:ecl}. We use the different values of $\rho$ to generate differently augmented versions of a clip. That way contrastive loss is  encouraged to learn spatial features which are invariant to such granularity changes. Apart from the temporal window, we also use a random event drop to get the differently augmented version of the same video instance.

\section{Experiments}
\subsection{Datasets}

\noindent \textbf{N-EPIC Kitchens}~\cite{nepic} is extended from EPIC-Kitchens \cite{Damen_2018_ECCV} to validate the performance of event cameras for ego-centric action recognition. Videos in the EPIC-Kitchens are first upsampled using Super SloMo \cite{jiang2018super} to match the micro-second temporal resolution of event cameras. Next, the event camera simulator ESIM \cite {rebecq2018esim} is used to convert the input RGB videos to event streams. Three largest kitchens from EPIC-Kitchens are converted named D1, D2, and D3. The dataset covers 8 action classes including \texttt{open}, \texttt{close}, \texttt{pour}, \texttt{mix}, \texttt{cut}, \texttt{take}, \texttt{wash}, and \texttt{put}. 

\noindent \textbf{DVS Gesture Recognition}~\cite{dvs} is collected using DVS128 camera with 128x128-pixel Dynamic Vision Sensor. The sensor can support a maximum event rate of 1M events/sec and provides a high dynamic range of 120 dB.  A total of 1,342 instances from 29 subjects are collected under 3 different lighting conditions. And, the dataset is divided into 11 hand and arm gestures including \texttt{hand waving}, \texttt{air drums}, \texttt{air guitar}, and \texttt{forearm rolling}. Following the standard setting \cite{dvs}, 23 subjects are used for training and the remaining 6 subjects are used for testing.

\subsection{Implementation Details}
\noindent\textbf{Architectural Detail} For VTN implementation, we use the original codebase\footnote{\href{https://github.com/bomri/SlowFast/tree/366467aafc856712fdc3e9c4cce8e90969047ee6/projects/vtn}{https://github.com/bomri/SlowFast}}.  For 2D-image encoder $F(\cdot)$, we show results on ViT-B~\cite{vit}. For Temporal Encoder $E(\cdot)$, we utilize a LongFormer of 3 layers with 8 attention heads. For the non-linear projection head of $\mathcal{L}_{EC}$, we utilize MLP of 1 hidden layer with output representation dimension = 128. 

\noindent\textbf{Input} We utilize spatial resolution of $224 \times 224$ with clip-length ($n$)= $16$.

\noindent\textbf{Training setup}  We train for $100$ epochs. For weight updates, we utilize Adam optimizer~\cite{adam} with $\beta_1=0.9, \beta_2=0.999, \epsilon=1e\minus8$,
the default parameters for PyTorch. We utilize a base learning rate of $4e\minus5$ with a linear warm-up of 10 epochs which is followed by a cosine learning scheduler for 90 epochs. For both losses $\mathcal{L}_{C}$ and $\mathcal{L}_{EC}$, we use equal weight i.e. $\alpha =1$. For contrastive loss $\mathcal{L}_{EC}$, we use temperature $\tau=0.1$.

\noindent\textbf{Inference setup} For inference, we adhere to the standard protocol~\cite{r2plus1d} of averaging the forecasts of 5 clips that are uniformly spaced, resulting in a prediction at the video-level.

\subsection{Results}
We assess our model's accuracy on two protocols for the DVS dataset: 10-class classification without the background class, and 11-class classification with the background class included.  As shown in Table~\ref{table:dvs}, our method achieves comparable results with previous studies. Note that while some studies like \cite{innocenti2021temporal} report similar performance using specialized image-encoding methods for DVS event data, our method does not utilize such specialized encoding for the DVS dataset.

\begin{table}
\vspace{2mm}
\centering
\caption{Comparison of gesture recognition performance on DVS dataset~\cite{dvs}.}
\label{table:dvs}

\begin{tabular}{lrr}
\hline

\hline

\hline\\[-3mm]
\textbf{Method}          & \textbf{10 classes} & \textbf{11 classes}  \\
\hline
Time-surfaces \cite{maro2020event}             & 96.59               & 90.62                \\
SNN eRBP \cite{Kaiser2019EmbodiedNV}           & -                   & 92.70                 \\
Slayer \cite{Shrestha2018SLAYERSL}             & -                   & 93.64                \\
Space-time clouds \cite{Wang2019SpaceTimeEC}   & 97.08               & 95.32                \\
DECOLLE \cite{Kaiser2018SynapticPD}            & -                   & 95.54                \\
Spatiotemporal filt. \cite{Ghosh2019SpatiotemporalFF} & -                   & 97.75                \\
RG-CNN \cite{Bi2019GraphBasedSF}              & -                   & 97.20                 \\
TBR \cite{innocenti2021temporal}                     & 97.50                & 97.73                \\
\textbf{Ours}                     & \textbf{97.69}                    & \textbf{97.92}                \\
\hline

\hline

\hline\\[-3mm]
\end{tabular}

\vspace{-3mm}
\end{table}

We also compare our method on N-EPIC Kitchens on two protocols: (1) Recognizing actions in the same kitchen, and (2) Recognizing actions in different kitchens. We compare various prior baselines mentioned in ~\cite{nepic}, which are trained with various other corresponding modalities with Events like RGB, and Optical-Flow. As shown in Table~\ref{table:nek}, our method outperforms all methods only using the single modality on both protocols. 

Our experimental results demonstrate the effectiveness of the proposed method for both gesture and real-world action recognition tasks. Particularly, the state-of-the-art performance on the N-EPIC Kitchens dataset, where the model is evaluated on unseen environments, highlights its ability to generalize to novel settings, making it a more reliable choice for deployment.

\subsection{Inference Accuracy vs Speed} 
\begin{figure*}[t]
    \centering
    \begin{subfigure}{0.24\linewidth}
        \vspace{2mm}
        \centering
        \includegraphics[width=0.97\linewidth]{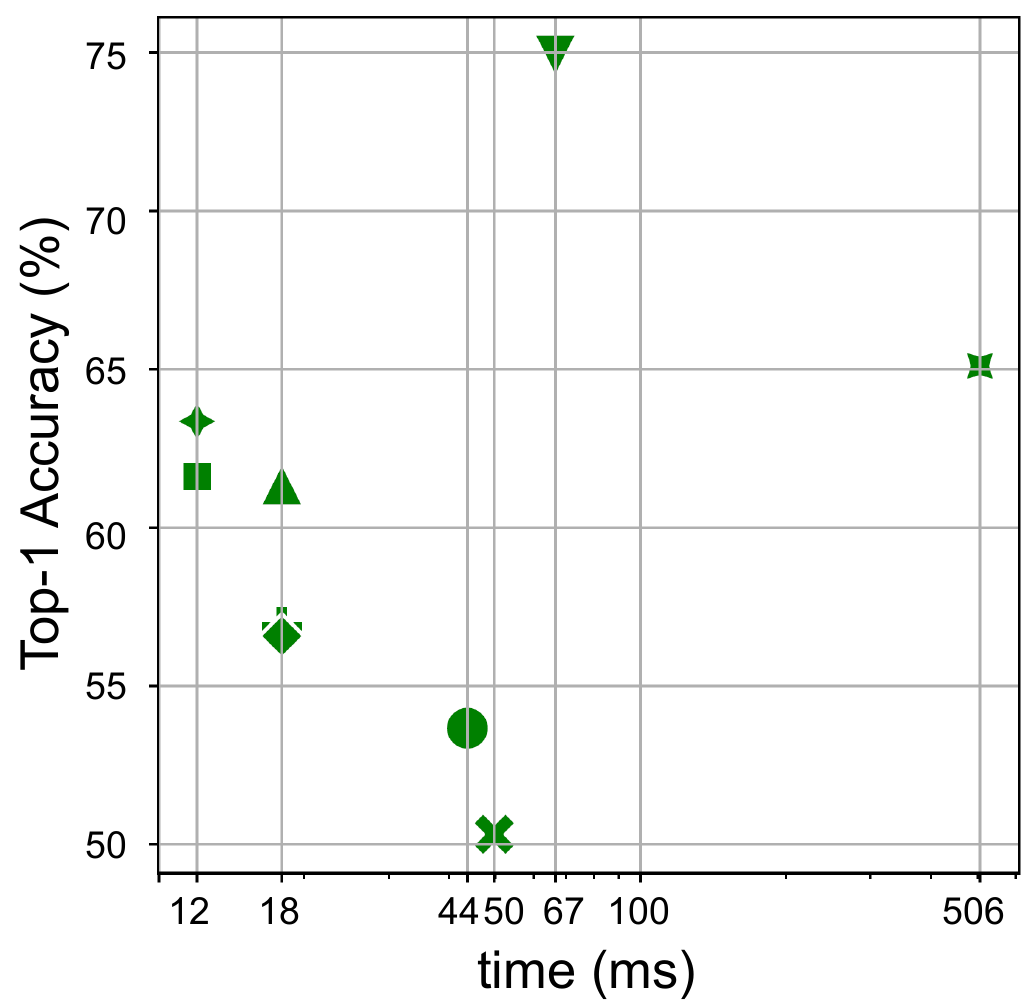}
        \caption{{Seen Kitchen}}

    \end{subfigure}
    \begin{subfigure}{0.24\linewidth}
        \vspace{2mm}
        \centering
        \includegraphics[clip, trim=0mm 0mm 0mm 5mm,width=\linewidth]{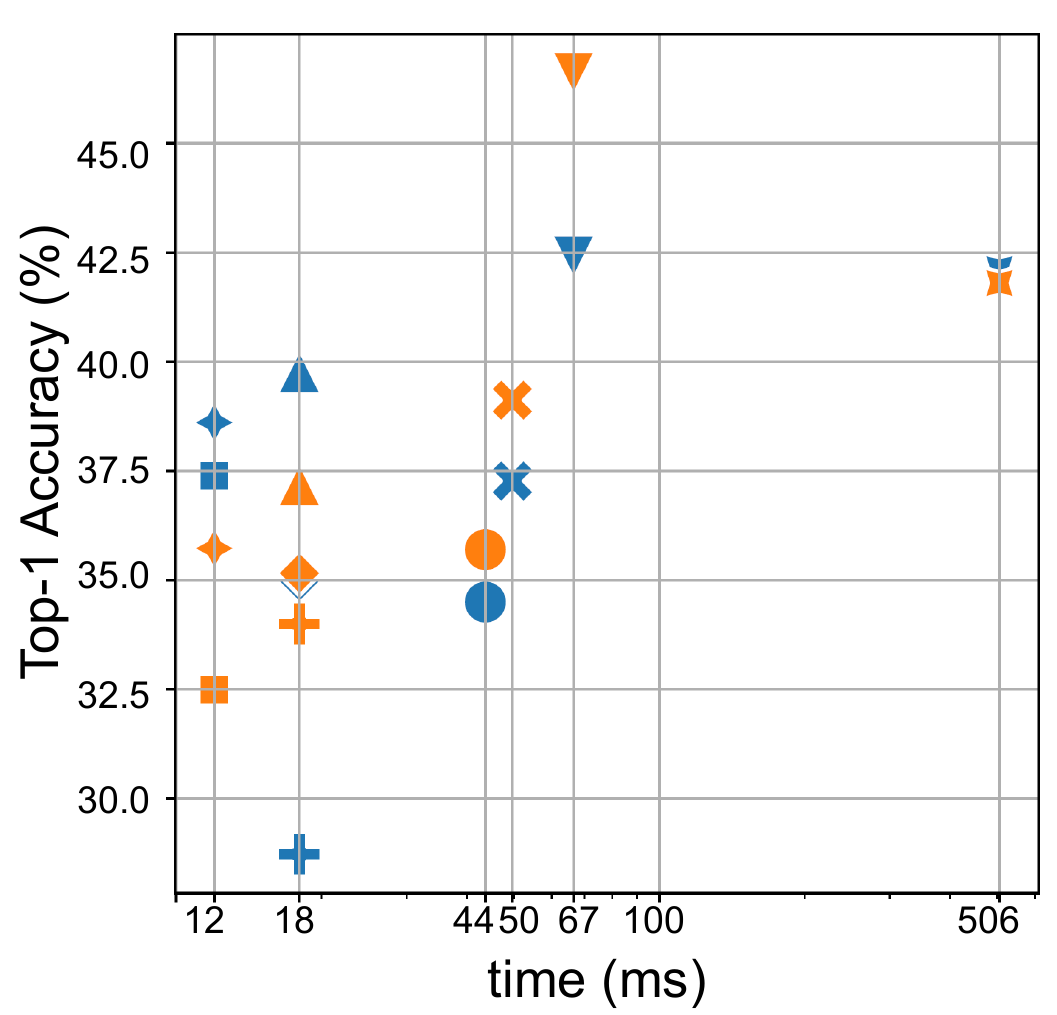}
        \caption{{Unseen Kitchen} }

    \end{subfigure}
    \begin{subfigure}{0.15\linewidth}
        \centering
        \includegraphics[width=0.97\linewidth]{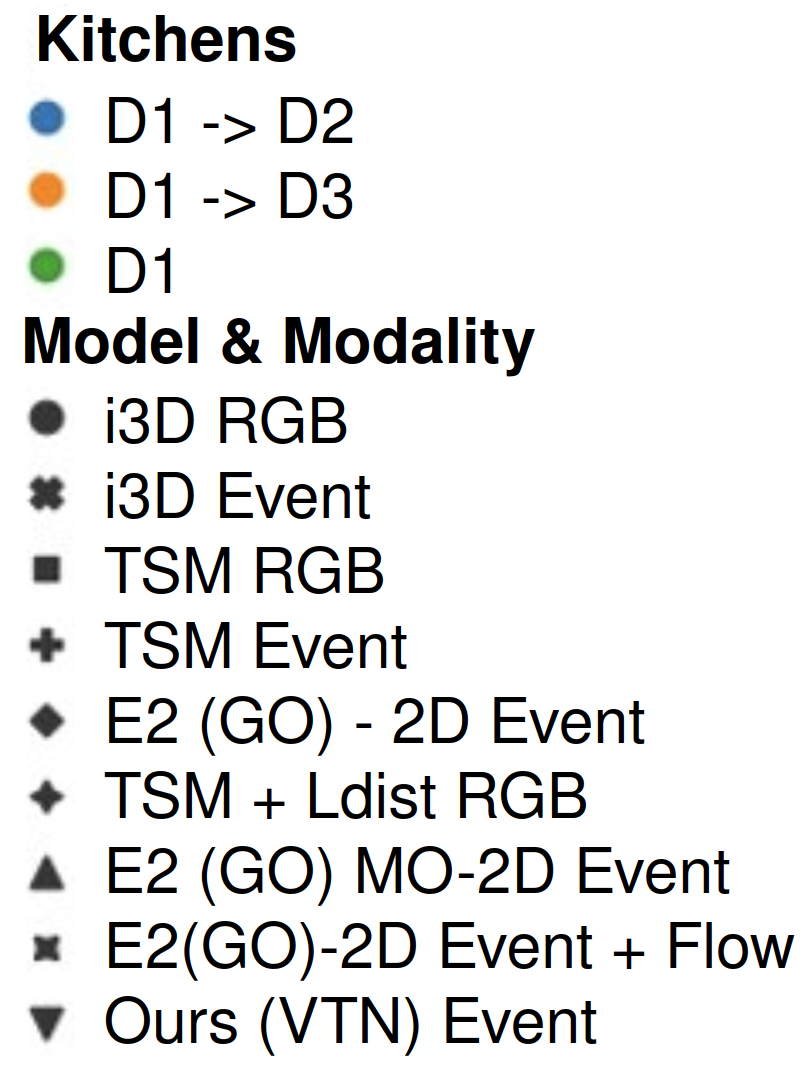}
        \vspace{4mm}

    \end{subfigure}

    \caption{\textbf{Accuracy vs Speed (Log-Scale) Trade-off:} The graph shows the trade-off between accuracy and speed for two settings of the N-EPIC Kitchens dataset, following the same protocol as in \cite{nepic}. Our proposed method using only event modality achieves better performance than previous methods, including the multi-modal approach of event+optical flow.}
    \label{fig:timing}
    \vspace{-3mm}
\end{figure*}

To assess the performance of our method in terms of inference time and compare it with prior work, we follow the timing protocol used by the prior state-of-the-art method~\cite{nepic}. We utilize an NVIDIA Titan RTX GPU for computation and report the input preprocessing time and forward time of a single clip (batch size = 1). 
We present the accuracy-time trade-off plot in Fig~\ref{fig:timing}, which illustrates the relationship between the model's performance and inference time (in milliseconds). Our approach achieves better accuracy than previous works in a comparable processing time and significantly outperforms the multimodal approach~\cite{nepic} in terms of both accuracy and computational speed.

\begin{table}
\vspace{2mm}
\centering
\caption{Comparison of action recognition performance on N-EPIC Kitchens dataset~\cite{nepic}.}
\label{table:nek}
\begingroup
\setlength{\tabcolsep}{4pt}
\begin{tabular}{lllccc}
\hline

\hline

\hline\\[-3mm]
{}  &\textbf{Model}       & \textbf{Modality}     & \textbf{D1}    & \textbf{D1$\rightarrow$D2} & \textbf{D1$\rightarrow$D3}  \\
\hline
\texttt{(a)}& I3D~\cite{kinetics}         & RGB          & 53.67 & 34.50   & 35.70   \\
\texttt{(b)}& I3D~\cite{kinetics}         & Event        & 50.32 & 37.27  & 39.12  \\
\texttt{(c)}& E2(GO)-3D~\cite{nepic}   & Event        & 50.52 & 38.07  & 38.71  \\
\texttt{(d)}& TSM~\cite{lin2019tsm}         & RGB          & 61.61 & 37.39  & 32.49  \\
\texttt{(e)}& TSM~\cite{lin2019tsm}         & Event        & 56.86 & 28.73  & 34.00     \\
\texttt{(f)}& E2(GO)-2D~\cite{nepic}   & Event        & 56.58 & 34.98  & 35.16  \\
\texttt{(g)}& TSM + Ldist         & RGB  & 63.36 & 38.61  & 35.73  \\
\texttt{(h)}& E2(GO)MO-2D~\cite{nepic} & Event        & 61.38 & 39.77  & 37.19  \\
\texttt{(i)}& E2(GO)-2D~\cite{nepic}   & Event+Flow & 65.11 & 42.12  & 41.80   \\
\texttt{(j)}& \textbf{Ours (VTN)}  & Event        & \textbf{74.90}  & \textbf{42.43}   & \textbf{46.66}  
  \\
\hline

\hline

\hline\\[-3mm]
\end{tabular}
\endgroup
\vspace{-3mm}
\end{table}

\subsection{Ablations}
\begin{table}[h]
\centering

\caption{Ablation with different Components on DVS}
\label{table:ablation}
\begin{tabular}{ll}
\hline

\hline

\hline\\[-3mm]
\textbf{Method}                                                                                               & \multicolumn{1}{l}{\textbf{Top-1 Accuracy (\%)}}  \\
\hline
Our Complete Framework                                                                                            & 
\textbf{97.92}                                    \\
Without Event Contrastive loss                                                                                      & 95.49 (-2.43)                                     \\
\begin{tabular}[c]{@{}l@{}}Without Event specific Augmentations \\(Time window $\rho$ and event drop)\end{tabular} & 92.71 (-5.21)  
\\
\hline

\hline

\hline\\[-3mm]
\end{tabular}

\end{table}

\noindent \textbf{Effect of Event-based augmentations and $\mathcal{L}_{EC}$} 
To investigate the impact of each component in our framework, we conduct an ablation study on the DVS dataset using all classes, as presented in Table~\ref{table:ablation}. The results indicate that removing the proposed event contrastive loss ($\mathcal{L}_{EC}$) causes a decrease in accuracy by 2.43\%, highlighting the importance of maintaining the fine-grained nature of the spatial embedding of VTN. Moreover, removing the proposed event-specific augmentations has the greatest negative impact on performance, mainly due to two reasons: (1) $\mathcal{L}_{EC}$ becomes trivial without event-specific augmentations, and (2) a decrease in augmentations leads to a smaller training set and potential overfitting.

\begin{table}
\vspace{2mm}
\centering
\caption{Ablation with different backbone on DVS}
\label{table:backbone}
\begin{tabular}{lrr}\hline

\hline

\hline\\[-3mm]
          & \multicolumn{1}{l}{\#parameters (M)} & \multicolumn{1}{l}{Top-1 Accuracy}  \\
          \hline
I3D~\cite{kinetics}       & 25.0                                   & 84.4                                \\
VideoSWIN~\cite{videoswin} & 88.1                                 & 95.7                                \\
VTN~\cite{vtn}       & 11.0                               & \textbf{97.9}                               \\
\hline

\hline

\hline\\[-3mm]
\end{tabular}
\vspace{-4mm}
\end{table}

\noindent \textbf{Effect of different video models} We conducted experiments using various video models such as I3D~\cite{kinetics}, VideoSWIN~\cite{videoswin}, and VTN~\cite{vtn}. Based on the results presented in Table~\ref{table:backbone}, it can be observed that the VTN architecture performs significantly better than the other models and requires minimum number of learnable parameters.

\section{Conclusion}
In this paper, we proposed a video transformer-based framework for event-camera based action recognition, which leverages event-contrastive loss and augmentations to adapt the network to event data. Our method achieved state-of-the-art results on N-EPIC Kitchens dataset and competitive results on standard DVS Gesture recognition datase, while requiring less computation time compared to competitive prior approaches. Our findings demonstrate the effectiveness of our proposed framework and highlight its potential impact in real-world applications.

Future research in this area could extend our work to Action Quality Assessment tasks that require more fine-grained temporal understanding than Action Recognition task, making them more relevant for event-based cameras. Another interesting direction could be exploring recent masked image modeling-based learning techniques to efficiently adapt RGB models to event data.
Overall, we believe that our work contributes to the advancement of event-based video understanding and provides a strong foundation for future research in this area.

{\bibliographystyle{IEEEtran}
\bibliography{main}

\begin{thebibliography}{10}
\providecommand{\url}[1]{#1}
\csname url@rmstyle\endcsname
\providecommand{\newblock}{\relax}
\providecommand{\bibinfo}[2]{#2}
\providecommand\BIBentrySTDinterwordspacing{\spaceskip=0pt\relax}
\providecommand\BIBentryALTinterwordstretchfactor{4}
\providecommand\BIBentryALTinterwordspacing{\spaceskip=\fontdimen2\font plus
\BIBentryALTinterwordstretchfactor\fontdimen3\font minus
  \fontdimen4\font\relax}
\providecommand\BIBforeignlanguage[2]{{%
\expandafter\ifx\csname l@#1\endcsname\relax
\typeout{** WARNING: IEEEtran.bst: No hyphenation pattern has been}%
\typeout{** loaded for the language `#1'. Using the pattern for}%
\typeout{** the default language instead.}%
\else
\language=\csname l@#1\endcsname
\fi
#2}}

\bibitem{gberta_2021_ICML}
G.~Bertasius, H.~Wang, and L.~Torresani, ``Is space-time attention all you need
  for video understanding?'' in \emph{Proceedings of the International
  Conference on Machine Learning (ICML)}, July 2021.

\bibitem{aim_chen}
T.~Yang, Y.~Zhu, Y.~Xie, A.~Zhang, C.~Chen, and M.~Li, ``Aim: Adapting image
  models for efficient video understanding,'' in \emph{International Conference
  on Learning Representations}, 2023.

\bibitem{videomae}
Z.~Tong, Y.~Song, J.~Wang, and L.~Wang, ``Video{MAE}: Masked autoencoders are
  data-efficient learners for self-supervised video pre-training,'' in
  \emph{Advances in Neural Information Processing Systems}, 2022.

\bibitem{gabv2}
I.~Dave, Z.~Scheffer, A.~Kumar, S.~Shiraz, Y.~S. Rawat, and M.~Shah,
  ``Gabriellav2: Towards better generalization in surveillance videos for
  action detection,'' in \emph{Proceedings of the IEEE/CVF Winter Conference on
  Applications of Computer Vision (WACV) Workshops}, January 2022, pp.
  122--132.

\bibitem{rizve2021gabriella}
M.~N. Rizve, U.~Demir, P.~Tirupattur, A.~J. Rana, K.~Duarte, I.~R. Dave, Y.~S.
  Rawat, and M.~Shah, ``Gabriella: An online system for real-time activity
  detection in untrimmed security videos,'' in \emph{2020 25th International
  Conference on Pattern Recognition (ICPR)}.\hskip 1em plus 0.5em minus
  0.4em\relax IEEE, 2021, pp. 4237--4244.

\bibitem{hvu}
A.~Diba, M.~Fayyaz, V.~Sharma, M.~Paluri, J.~Gall, R.~Stiefelhagen, and
  L.~Van~Gool, ``Large scale holistic video understanding,'' in \emph{European
  Conference on Computer Vision}.\hskip 1em plus 0.5em minus 0.4em\relax
  Springer, 2020, pp. 593--610.

\bibitem{kinetics}
J.~Carreira and A.~Zisserman, ``Quo vadis, action recognition? a new model and
  the kinetics dataset,'' in \emph{proceedings of the IEEE Conference on
  Computer Vision and Pattern Recognition}, 2017, pp. 6299--6308.

\bibitem{hacs}
H.~Zhao, A.~Torralba, L.~Torresani, and Z.~Yan, ``Hacs: Human action clips and
  segments dataset for recognition and temporal localization,'' in
  \emph{Proceedings of the IEEE International Conference on Computer Vision},
  2019, pp. 8668--8678.

\bibitem{miech2019howto100m}
A.~Miech, D.~Zhukov, J.-B. Alayrac, M.~Tapaswi, I.~Laptev, and J.~Sivic,
  ``Howto100m: Learning a text-video embedding by watching hundred million
  narrated video clips,'' in \emph{Proceedings of the IEEE/CVF International
  Conference on Computer Vision}, 2019, pp. 2630--2640.

\bibitem{schiappa2022large}
M.~C. Schiappa, N.~Biyani, S.~Vyas, H.~Palangi, V.~Vineet, and Y.~Rawat,
  ``Large-scale robustness analysis of video action recognition models,''
  \emph{arXiv preprint arXiv:2207.01398}, 2022.

\bibitem{gallego2020event}
G.~Gallego, T.~Delbr{\"u}ck, G.~Orchard, C.~Bartolozzi, B.~Taba, A.~Censi,
  S.~Leutenegger, A.~J. Davison, J.~Conradt, K.~Daniilidis, \emph{et~al.},
  ``Event-based vision: A survey,'' \emph{IEEE transactions on pattern analysis
  and machine intelligence}, vol.~44, no.~1, pp. 154--180, 2020.

\bibitem{ahmad2022event}
S.~Ahmad, G.~Scarpellini, P.~Morerio, and A.~Del~Bue, ``Event-driven re-id: a
  new benchmark and method towards privacy-preserving person
  re-identification,'' in \emph{Proceedings of the IEEE/CVF Winter Conference
  on Applications of Computer Vision}, 2022, pp. 459--468.

\bibitem{dave2022spact}
I.~R. Dave, C.~Chen, and M.~Shah, ``Spact: Self-supervised privacy preservation
  for action recognition,'' in \emph{Proceedings of the IEEE/CVF Conference on
  Computer Vision and Pattern Recognition}, 2022, pp. 20\,164--20\,173.

\bibitem{fioresi2023tedspad}
J.~Fioresi, I.~Dave, and M.~Shah, ``Ted-spad: Temporal distinctiveness for
  self-supervised privacy-preservation for video anomaly detection,'' in
  \emph{ICCV}, 2023.

\bibitem{nepic}
C.~Plizzari, M.~Planamente, G.~Goletto, M.~Cannici, E.~Gusso, M.~Matteucci, and
  B.~Caputo, ``E2 (go) motion: Motion augmented event stream for egocentric
  action recognition,'' in \emph{Proceedings of the IEEE/CVF Conference on
  Computer Vision and Pattern Recognition}, 2022, pp. 19\,935--19\,947.

\bibitem{dvs}
A.~Amir, B.~Taba, D.~Berg, T.~Melano, J.~McKinstry, C.~Di~Nolfo, T.~Nayak,
  A.~Andreopoulos, G.~Garreau, M.~Mendoza, \emph{et~al.}, ``A low power, fully
  event-based gesture recognition system,'' in \emph{Proceedings of the IEEE
  conference on computer vision and pattern recognition}, 2017, pp. 7243--7252.

\bibitem{vtn}
D.~Neimark, O.~Bar, M.~Zohar, and D.~Asselmann, ``Video transformer network,''
  in \emph{Proceedings of the IEEE/CVF International Conference on Computer
  Vision (ICCV) Workshops}, October 2021, pp. 3163--3172.

\bibitem{zhang2021vidtr}
Y.~Zhang, X.~Li, C.~Liu, B.~Shuai, Y.~Zhu, B.~Brattoli, H.~Chen, I.~Marsic, and
  J.~Tighe, ``Vidtr: Video transformer without convolutions,'' in
  \emph{Proceedings of the IEEE/CVF international conference on computer
  vision}, 2021, pp. 13\,577--13\,587.

\bibitem{hochreiter1997long}
S.~Hochreiter and J.~Schmidhuber, ``Long short-term memory,'' \emph{Neural
  computation}, vol.~9, no.~8, pp. 1735--1780, 1997.

\bibitem{Jiao_2021_CVPR}
J.~Jiao, H.~Huang, L.~Li, Z.~He, Y.~Zhu, and M.~Liu, ``Comparing
  representations in tracking for event camera-based slam,'' in
  \emph{Proceedings of the IEEE/CVF Conference on Computer Vision and Pattern
  Recognition (CVPR) Workshops}, June 2021, pp. 1369--1376.

\bibitem{perot2020learning}
E.~Perot, P.~De~Tournemire, D.~Nitti, J.~Masci, and A.~Sironi, ``Learning to
  detect objects with a 1 megapixel event camera,'' \emph{Advances in Neural
  Information Processing Systems}, vol.~33, pp. 16\,639--16\,652, 2020.

\bibitem{8880496}
M.~Almatrafi and K.~Hirakawa, ``Davis camera optical flow,'' \emph{IEEE
  Transactions on Computational Imaging}, vol.~6, pp. 396--407, 2020.

\bibitem{stoffregen2020reducing}
T.~Stoffregen, C.~Scheerlinck, D.~Scaramuzza, T.~Drummond, N.~Barnes,
  L.~Kleeman, and R.~Mahony, ``Reducing the sim-to-real gap for event
  cameras,'' in \emph{Computer Vision--ECCV 2020: 16th European Conference,
  Glasgow, UK, August 23--28, 2020, Proceedings, Part XXVII 16}.\hskip 1em plus
  0.5em minus 0.4em\relax Springer, 2020, pp. 534--549.

\bibitem{kim2021n}
J.~Kim, J.~Bae, G.~Park, D.~Zhang, and Y.~M. Kim, ``N-imagenet: Towards robust,
  fine-grained object recognition with event cameras,'' in \emph{Proceedings of
  the IEEE/CVF International Conference on Computer Vision}, 2021, pp.
  2146--2156.

\bibitem{alonso2019ev}
I.~Alonso and A.~C. Murillo, ``Ev-segnet: Semantic segmentation for event-based
  cameras,'' in \emph{Proceedings of the IEEE/CVF Conference on Computer Vision
  and Pattern Recognition Workshops}, 2019, pp. 0--0.

\bibitem{chen2022ecsnet}
Z.~Chen, J.~Wu, J.~Hou, L.~Li, W.~Dong, and G.~Shi, ``Ecsnet: Spatio-temporal
  feature learning for event camera,'' \emph{IEEE Transactions on Circuits and
  Systems for Video Technology}, 2022.

\bibitem{lu2020event}
J.~Lu, J.~Dong, R.~Yan, and H.~Tang, ``An event-based categorization model
  using spatio-temporal features in a spiking neural network,'' in \emph{2020
  12th International Conference on Advanced Computational Intelligence
  (ICACI)}.\hskip 1em plus 0.5em minus 0.4em\relax IEEE, 2020, pp. 385--390.

\bibitem{Zhu_2019_CVPR}
A.~Z. Zhu, L.~Yuan, K.~Chaney, and K.~Daniilidis, ``Unsupervised event-based
  learning of optical flow, depth, and egomotion,'' in \emph{Proceedings of the
  IEEE/CVF Conference on Computer Vision and Pattern Recognition (CVPR)}, June
  2019.

\bibitem{rebecq2019events}
H.~Rebecq, R.~Ranftl, V.~Koltun, and D.~Scaramuzza, ``Events-to-video: Bringing
  modern computer vision to event cameras,'' in \emph{Proceedings of the
  IEEE/CVF Conference on Computer Vision and Pattern Recognition}, 2019, pp.
  3857--3866.

\bibitem{ulhaq2022vision}
A.~Ulhaq, N.~Akhtar, G.~Pogrebna, and A.~Mian, ``Vision transformers for action
  recognition: A survey,'' \emph{arXiv preprint arXiv:2209.05700}, 2022.

\bibitem{sun2022human}
Z.~Sun, Q.~Ke, H.~Rahmani, M.~Bennamoun, G.~Wang, and J.~Liu, ``Human action
  recognition from various data modalities: A review,'' \emph{IEEE transactions
  on pattern analysis and machine intelligence}, 2022.

\bibitem{simclr}
T.~Chen, S.~Kornblith, M.~Norouzi, and G.~Hinton, ``A simple framework for
  contrastive learning of visual representations,'' in \emph{ICML}, 2020.

\bibitem{moco}
K.~He, H.~Fan, Y.~Wu, S.~Xie, and R.~Girshick, ``Momentum contrast for
  unsupervised visual representation learning,'' in \emph{Proceedings of the
  IEEE/CVF Conference on Computer Vision and Pattern Recognition}, 2020, pp.
  9729--9738.

\bibitem{swav}
M.~Caron, I.~Misra, J.~Mairal, P.~Goyal, P.~Bojanowski, and A.~Joulin,
  ``Unsupervised learning of visual features by contrasting cluster
  assignments,'' in \emph{Advances in Neural Information Processing Systems},
  H.~Larochelle, M.~Ranzato, R.~Hadsell, M.~F. Balcan, and H.~Lin, Eds.,
  vol.~33.\hskip 1em plus 0.5em minus 0.4em\relax Curran Associates, Inc.,
  2020, pp. 9912--9924.

\bibitem{wu2018unsupervised}
Z.~Wu, Y.~Xiong, S.~X. Yu, and D.~Lin, ``Unsupervised feature learning via
  non-parametric instance discrimination,'' in \emph{Proceedings of the IEEE
  Conference on Computer Vision and Pattern Recognition}, 2018, pp. 3733--3742.

\bibitem{tclr}
I.~Dave, R.~Gupta, M.~N. Rizve, and M.~Shah, ``Tclr: Temporal contrastive
  learning for video representation,'' \emph{Computer Vision and Image
  Understanding}, p. 103406, 2022.

\bibitem{timebalance}
I.~R. Dave, M.~N. Rizve, C.~Chen, and M.~Shah, ``Timebalance:
  Temporally-invariant and temporally-distinctive video representations for
  semi-supervised action recognition,'' in \emph{Proceedings of the IEEE/CVF
  Conference on Computer Vision and Pattern Recognition}, 2023.

\bibitem{c3d}
D.~Tran, L.~Bourdev, R.~Fergus, L.~Torresani, and M.~Paluri, ``Learning
  spatiotemporal features with 3d convolutional networks,'' in
  \emph{Proceedings of the IEEE international conference on computer vision},
  2015, pp. 4489--4497.

\bibitem{feichtenhofer2020x3d}
C.~Feichtenhofer, ``X3d: Expanding architectures for efficient video
  recognition,'' in \emph{Proceedings of the IEEE/CVF Conference on Computer
  Vision and Pattern Recognition}, 2020, pp. 203--213.

\bibitem{kenshohara}
K.~{Hara}, H.~{Kataoka}, and Y.~{Satoh}, ``Towards good practice for action
  recognition with spatiotemporal 3d convolutions,'' in \emph{2018 24th
  International Conference on Pattern Recognition (ICPR)}, 2018, pp.
  2516--2521.

\bibitem{vaswani2017attention}
A.~Vaswani, N.~Shazeer, N.~Parmar, J.~Uszkoreit, L.~Jones, A.~N. Gomez,
  {\L}.~Kaiser, and I.~Polosukhin, ``Attention is all you need,''
  \emph{Advances in neural information processing systems}, vol.~30, 2017.

\bibitem{mocov3}
X.~Chen, S.~Xie, and K.~He, ``An empirical study of training self-supervised
  vision transformers,'' in \emph{Proceedings of the IEEE/CVF International
  Conference on Computer Vision}, 2021, pp. 9640--9649.

\bibitem{Damen_2018_ECCV}
D.~Damen, H.~Doughty, G.~M. Farinella, S.~Fidler, A.~Furnari, E.~Kazakos,
  D.~Moltisanti, J.~Munro, T.~Perrett, W.~Price, and M.~Wray, ``Scaling
  egocentric vision: The epic-kitchens dataset,'' in \emph{Proceedings of the
  European Conference on Computer Vision (ECCV)}, September 2018.

\bibitem{jiang2018super}
H.~Jiang, D.~Sun, V.~Jampani, M.-H. Yang, E.~Learned-Miller, and J.~Kautz,
  ``Super slomo: High quality estimation of multiple intermediate frames for
  video interpolation,'' in \emph{Proceedings of the IEEE conference on
  computer vision and pattern recognition}, 2018, pp. 9000--9008.

\bibitem{rebecq2018esim}
H.~Rebecq, D.~Gehrig, and D.~Scaramuzza, ``Esim: an open event camera
  simulator,'' in \emph{Conference on robot learning}.\hskip 1em plus 0.5em
  minus 0.4em\relax PMLR, 2018, pp. 969--982.

\bibitem{vit}
A.~Dosovitskiy, L.~Beyer, A.~Kolesnikov, D.~Weissenborn, X.~Zhai,
  T.~Unterthiner, M.~Dehghani, M.~Minderer, G.~Heigold, S.~Gelly,
  \emph{et~al.}, ``An image is worth 16x16 words: Transformers for image
  recognition at scale,'' \emph{arXiv preprint arXiv:2010.11929}, 2020.

\bibitem{adam}
D.~P. Kingma and J.~Ba, ``Adam: {A} method for stochastic optimization,'' in
  \emph{3rd International Conference on Learning Representations, {ICLR} 2015,
  San Diego, CA, USA, May 7-9, 2015, Conference Track Proceedings}, Y.~Bengio
  and Y.~LeCun, Eds., 2015.

\bibitem{r2plus1d}
D.~Tran, H.~Wang, L.~Torresani, J.~Ray, Y.~LeCun, and M.~Paluri, ``A closer
  look at spatiotemporal convolutions for action recognition,'' in
  \emph{Proceedings of the IEEE conference on Computer Vision and Pattern
  Recognition}, 2018, pp. 6450--6459.

\bibitem{innocenti2021temporal}
S.~U. Innocenti, F.~Becattini, F.~Pernici, and A.~Del~Bimbo, ``Temporal binary
  representation for event-based action recognition,'' in \emph{2020 25th
  International Conference on Pattern Recognition (ICPR)}.\hskip 1em plus 0.5em
  minus 0.4em\relax IEEE, 2021, pp. 10\,426--10\,432.

\bibitem{maro2020event}
J.-M. Maro, S.-H. Ieng, and R.~Benosman, ``Event-based gesture recognition with
  dynamic background suppression using smartphone computational capabilities,''
  \emph{Frontiers in neuroscience}, vol.~14, p. 275, 2020.

\bibitem{Kaiser2019EmbodiedNV}
J.~Kaiser, A.~Friedrich, J.~C.~V. Tieck, D.~Reichard, A.~Roennau, E.~O. Neftci,
  and R.~Dillmann, ``Embodied neuromorphic vision with event-driven random
  backpropagation,'' \emph{arXiv: Neural and Evolutionary Computing}, 2019.

\bibitem{Shrestha2018SLAYERSL}
S.~Shrestha and G.~Orchard, ``Slayer: Spike layer error reassignment in time,''
  \emph{ArXiv}, vol. abs/1810.08646, 2018.

\bibitem{Wang2019SpaceTimeEC}
Q.~Wang, Y.~Zhang, J.~Yuan, and Y.~Lu, ``Space-time event clouds for gesture
  recognition: From rgb cameras to event cameras,'' \emph{2019 IEEE Winter
  Conference on Applications of Computer Vision (WACV)}, pp. 1826--1835, 2019.

\bibitem{Kaiser2018SynapticPD}
J.~Kaiser, H.~Mostafa, and E.~O. Neftci, ``Synaptic plasticity dynamics for
  deep continuous local learning (decolle),'' \emph{Frontiers in Neuroscience},
  vol.~14, 2018.

\bibitem{Ghosh2019SpatiotemporalFF}
R.~Ghosh, A.~K. Gupta, A.~N. Silva, A.~B. Soares, and N.~V. Thakor,
  ``Spatiotemporal filtering for event-based action recognition,''
  \emph{ArXiv}, vol. abs/1903.07067, 2019.

\bibitem{Bi2019GraphBasedSF}
Y.~Bi, A.~Chadha, A.~Abbas, E.~Bourtsoulatze, and Y.~Andreopoulos,
  ``Graph-based spatio-temporal feature learning for neuromorphic vision
  sensing,'' \emph{IEEE Transactions on Image Processing}, vol.~29, pp.
  9084--9098, 2019.

\bibitem{lin2019tsm}
J.~Lin, C.~Gan, and S.~Han, ``Tsm: Temporal shift module for efficient video
  understanding,'' in \emph{Proceedings of the IEEE/CVF international
  conference on computer vision}, 2019, pp. 7083--7093.

\bibitem{videoswin}
Z.~Liu, J.~Ning, Y.~Cao, Y.~Wei, Z.~Zhang, S.~Lin, and H.~Hu, ``Video swin
  transformer,'' in \emph{Proceedings of the IEEE/CVF conference on computer
  vision and pattern recognition}, 2022, pp. 3202--3211.

\end{thebibliography}
}

\end{document}